\documentclass[lettersize,journal]{IEEEtran}
\usepackage{amsmath,amsfonts}
\usepackage{algorithmic}
\usepackage{algorithm}
\usepackage{array}
\usepackage[caption=false,font=normalsize,labelfont=sf,textfont=sf]{subfig}
\usepackage{textcomp}
\usepackage{stfloats}
\usepackage{url}
\usepackage{verbatim}
\usepackage{graphicx}
\usepackage{cite}
\usepackage{amsmath,amssymb,amsfonts}
\usepackage{algorithmic}
\usepackage{graphicx}
\usepackage{textcomp}
\usepackage{xcolor}
\usepackage{stfloats}
\usepackage{lipsum}
\usepackage{amsfonts}
\usepackage{multirow}
\usepackage{booktabs}
\usepackage{makecell}
\usepackage{doi}
\usepackage{pifont}
\usepackage{etoolbox}

\begin{document}

\title{MagicFace: High-Fidelity Facial Expression Editing with Action-Unit Control}

\author{Mengting Wei, 
      Tuomas Varanka,
      Xingxun Jiang,
      Huai-Qian Khor,
      Guoying Zhao$^{\ast}$,~\IEEEmembership{Fellow,~IEEE}

   \thanks{M. Wei, T. Varanka,  H. Khor and G. Zhao are with the Center for Machine Vision and Signal Analysis, Faculty of Information Technology and Electrical Engineering, University of Oulu, Oulu, FI-90014, Finland. E-mail: \{mengting.wei, chen.haoyu, yante.li, guoying.zhao\}@oulu.fi.}
   \thanks{X. Jiang is with the Key Laboratory of Child Development and Learning Science of Ministry of Education, School of Biological Sciences and Medical Engineering, Southeast University, Nanjing 210096, China, and is also with the Center for Machine Vision and Signal Analysis, Faulty of Information Technology and Electrical Engineering, University of Oulu, Oulu, FI-90014, Finland (e-mail:jiangxingxun@seu.edu.cn).}
   \thanks{*Corresponding author}
}

\markboth{Journal of \LaTeX\ Class Files,~Vol.~14, No.~8, August~2021}%
{Shell \MakeLowercase{\textit{et al.}}: A Sample Article Using IEEEtran.cls for IEEE Journals}

\IEEEpubid{0000--0000/00\$00.00~\copyright~2021 IEEE}

\maketitle

\begin{abstract}
We address the problem of facial expression editing by controlling the relative variation of facial action unit (AU) from the same person. This enables us to edit this specific person's expression in a fine-grained, continuous, and interpretable manner, while preserving their identity, pose, background, and detailed facial attributes. Key to our model, which we dub MagicFace, is a diffusion model conditioned on AU variations and an ID encoder to preserve facial details of high consistency. Specifically, to preserve the facial details with the input identity, we leverage the power of pretrained Stable-Diffusion models and design an ID encoder to merge appearance features through self-attention. To keep background and pose consistency, we introduce an efficient Attribute Controller by explicitly informing the model of the current background and pose of the target. By injecting AU variations into a denoising UNet, our model can animate arbitrary identities with various AU combinations, yielding superior results in high-fidelity expression editing compared to other facial expression editing works. Code is publicly available at https://github.com/weimengting/MagicFace.  
\end{abstract}


\section{Introduction}\label{sec:introd}

It is a perennial challenge in computer vision to realistically change the expression of a close-up while preserving the person's identity and other attributes, either from the background or other face characteristics. The challenge of this problem arises from the lack of an intuitive and interpretable depiction to represent facial expressions that can support customized expressions, and previous work usually addresses this with a latent space of expressions, where the codes are learned from a large expression dataset or the off-the-shelf ones like 3DMM parameters \cite{yan2019joint,zhu2024stableswap,liang2019adaptive,hou2022textface,peng2011editing}. These methods ignore the fact that the semantic meanings of these codes are implicit, which poses challenges for interpretable, arbitrary, and flexible manipulation of expression by non-professionals. In this work, we show that it's possible to convincingly alter a person's expression in a user-friendly manner by offering localized control with adjustable intensity while preserving their identity and other attributes from the portrait. Our key insight is to employ action units (AUs) to represent facial expressions, and then steer a Stable-Diffusion model to produce high-quality expression editing results.

\begin{figure*}[t] 
\centering 
\includegraphics[width=1.0\textwidth]{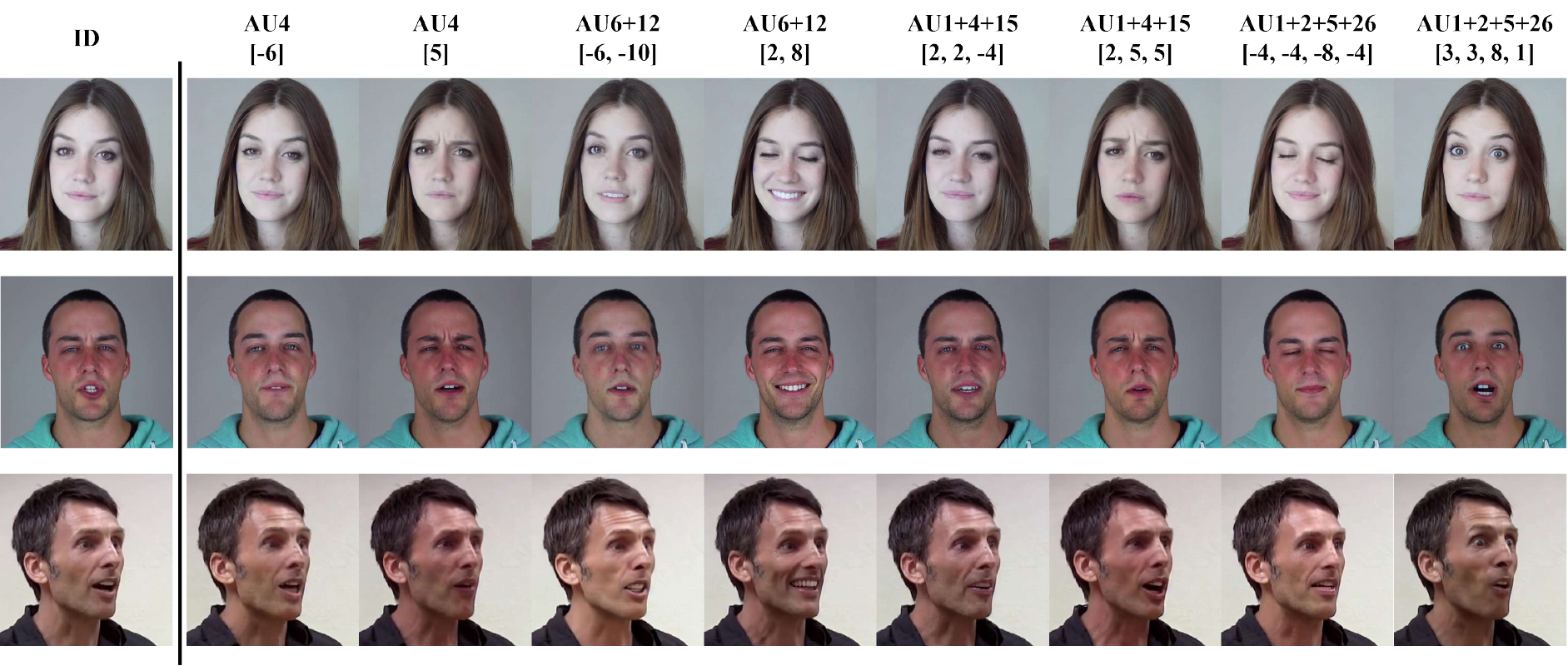}
\caption{MagicFace takes in the AU changes based on the input portrait and edit the portrait to exhibit different expressions. The edited image respects the AU condition and preserve identity, pose, background as well as other facial details.}
\label{fig:demo}
\end{figure*}

In editing facial expressions, the crucial question is which representation should be used to encode expressions. Facial AUs, as anatomical markers of facial muscle activity, have proven effective for precise and flexible facial expression description in images \cite{wang2019weakly,yan2022weakly,karaouglu2021self}. However, most AU annotations are of frontal faces of a limited number of subjects in laboratory settings, which easily to cause overfitting when used for training. A model trained by this type of data is hard to generalize to individuals with different poses and backgrounds, which we will present in Section \ref{sec:abla}. To deal with this issue, off-the-shelf AU intensity estimators like Libreface \cite{chang2024libreface} used a knowledge distillation technique from a large-scale network pre-trained on natural images to accommodate the AU intensity estimation task on the lab datasets. Such automatic tools provide us with AU intensity estimation of high accuracy, hence we can produce AU estimation of any face image and use the estimation as an AU condition to train our model.

\IEEEpubidadjcol

On the other hand, diffusion models have emerged as a popular choice for image generation, surpassing Generative Adversarial Networks (GANs) with higher generation quality \cite{zhang2023transformer,zhang2020supervised,xie2023consistency,yan2021fine}. ControlNet-style models enable users to add additional control signals, such as depth, paintings and skeleton pose, as the condition of the target \cite{zhang2023adding}. Some works modify it to enable some basic expressions like \textit{happiness} and \textit{sad} \cite{preechakul2021diffusion,zhang2023adding}, but many of their editing results are with extreme face deformations that may look unrealistic. More importantly, these models can not provide specific controls over expressions, whether through text or through images. Such specific controls, including intensity and location of expressions, are the focus of our work.

To merge the advantages of both worlds, we propose MagicFace, a model that allows us to correlate AU changes to facial expressions and then produce a photorealistic edited image conditioned on AU variations. Specifically, MagicFace first extracts AU intensities from portrait photographs using an off-the-shelf method, performs desired AU changes, and finally uses a Stable-Diffusion model to map the AU variations into photorealistic images. As the edited images should preserve the background and pose, we first cut out the background of the portrait and draw the contour of the pose, which is learned by an Attribute Controller, so that MagicFace only needs to perform conditional inpainting to the face. To maintain the consistency of identity and high-frequency facial characteristics, we introduce an ID encoder, which is a symmetrical UNet structure to capture spatial details of the input identity. This design allows the model to understand the relationship with the identity image within a uniform feature space, greatly enhancing the preservation of detailed appearance features. Our model is trained on a designed dataset with 30K image pairs. Fig. \ref{fig:demo} shows the editing results of various people, background, and AU combinations.

In summary, our contributions are:

\begin{itemize}
    \item A generative model that enables precise and localized facial expression editing by using AU variations to represent facial expressions.
    \item An ID encoder capable of well preserving the attributes of the person in the edited portraits, unconstrained by any head poses, backgrounds, and characters.  
    \item Quantitative and qualitative results show that our method presents a more interpretable, flexible, and user-friendly editing manner with higher image generation quality compared to other facial expression editing methods. 
\end{itemize}

\section{Related Works}

\begin{figure}[t] 
\centering 
\includegraphics[width=0.48\textwidth]{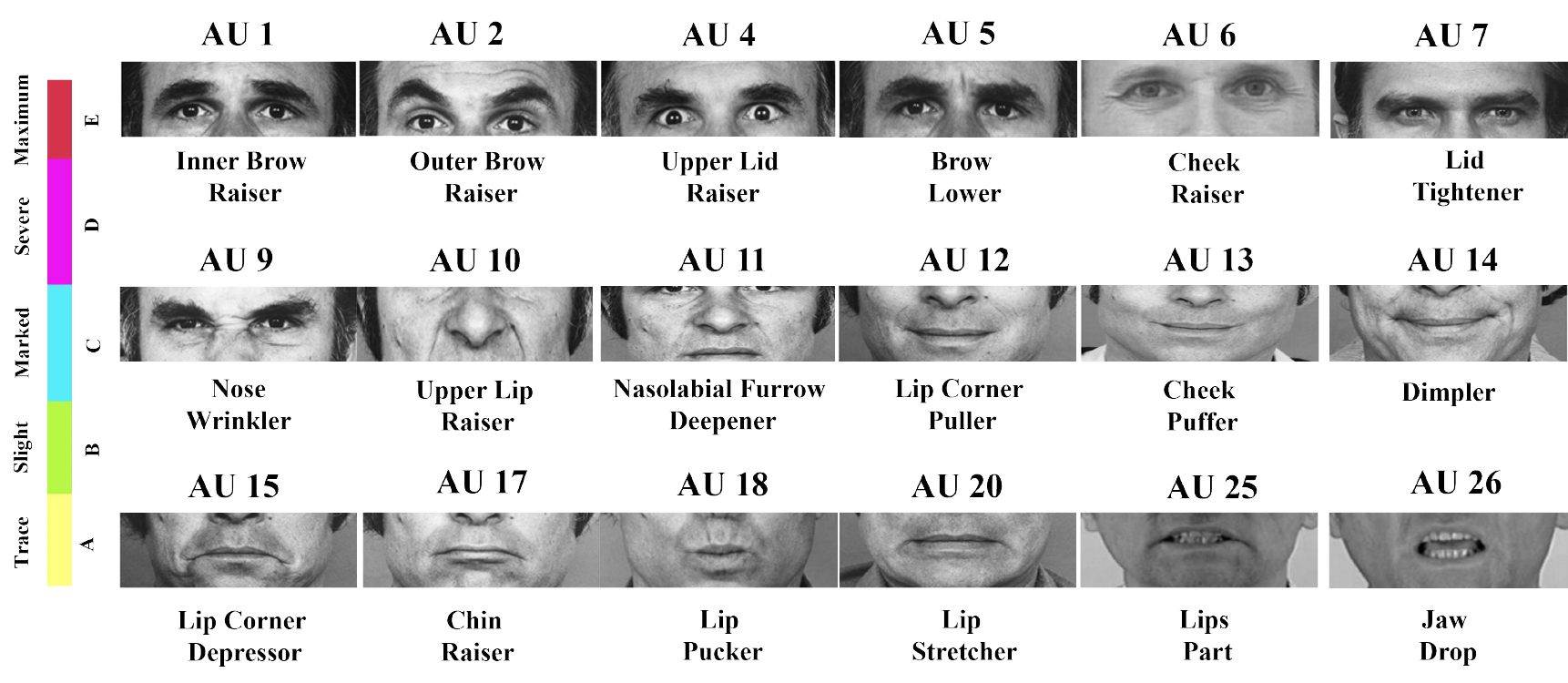}
\caption{\textbf{A display showcasing various action units and their corresponding intensity scales.} Only a set of commonly used AUs is displayed here. For a complete collection of AUs with descriptions, see \cite{ozelfaces}.}
\label{fig:au}
\end{figure}

\subsection{Facial Action Units}
The Facial Action Coding System (FACS) \cite{ekman1978facial} is one of the most impactful methods for analyzing facial behavior. It is a comprehensive, anatomy-based system capable of encoding diverse facial movements through combinations of fundamental Action Units (AUs). AUs represent specific facial configurations resulting from the contraction of one or more facial muscles and are not influenced by emotional interpretation. The earlier version of FACS \cite{ekman1978facial} included 44 Action Units (AUs), with 30 of them anatomically linked to specific facial muscles, while the remaining 14 were categorized as miscellaneous actions. In a later version \cite{ekman2002facial}, the criteria were updated: AU25, AU26, and AU27 were merged based on intensity, as were AU41, AU42, and AU43. The intensities of AUs can be assessed using a five-point ordinal scale, typically labeled with uppercase letters A to E, representing minimal to maximal intensity alongside the AU number. For example, AU2A indicates the minimum intensity of AU2, while AU2E represents its maximum intensity. However, this five-point scale is not uniform; for instance, levels C and D encompass a broader range of facial changes compared to levels A, B, and E.  Fig. \ref{fig:au} shows example images labeled with specific AUs. The precision and clarity of AUs provide unmatched control over facial expressions, offering users an intuitive and easily interpretable set of tools, so we choose to use it as the condition to describe and edit facial expressions.

\subsection{Facial Expression Editing}
Facial expression editing is a challenging task as it demands a deep understanding of input facial images and prior knowledge about human expressions. Unlike general facial attribute editing, which primarily focuses on modifying the appearance of specific facial regions, facial expression editing is more complex as it often involves significant geometric transformations and simultaneous changes across multiple facial components. Remarkable progress has been made in earlier years with the advancement of GANs. ExprGAN \cite{ding2018exprgan} introduces an expression controller to adjust the intensity of generated expressions, but it relies on a pre-trained face recognizer to maintain identity information. StarGAN \cite{choi2018stargan}, on the other hand, enables image translation across domains using a single model and preserves identity features by minimizing a cycle loss. However, it is limited to generating only discrete expressions. GANimation \cite{pumarola2018ganimation} provides a more fine-grained method for editing facial expressions by leveraging Action Units (AUs). Similarly, ICface \cite{tripathy2020icface} utilizes AUs to enable controllable facial expressions in facial reenactment. Recent focus has shifted to diffusion models due to their ability to produce high-quality images. However, many text-to-image diffusion models are unable to support flexible facial expression editing and, in most cases, can only handle simple emotion labels. Instead, some approaches choose the expression coefficients of 3DMM as the condition to enable fine-grained control over facial expression. However, the large number of coefficients poses a significant challenge, as only experts can manually adjust the desired expressions, which limits its applicability for general user-level applications. In contrast, this work focuses on injecting AUs into pre-trained large diffusion models to enable more flexible and user-friendly editing of the face.


\section{Method}

\begin{figure*}[htbp]
    \centering
    \includegraphics[width=0.96\textwidth]{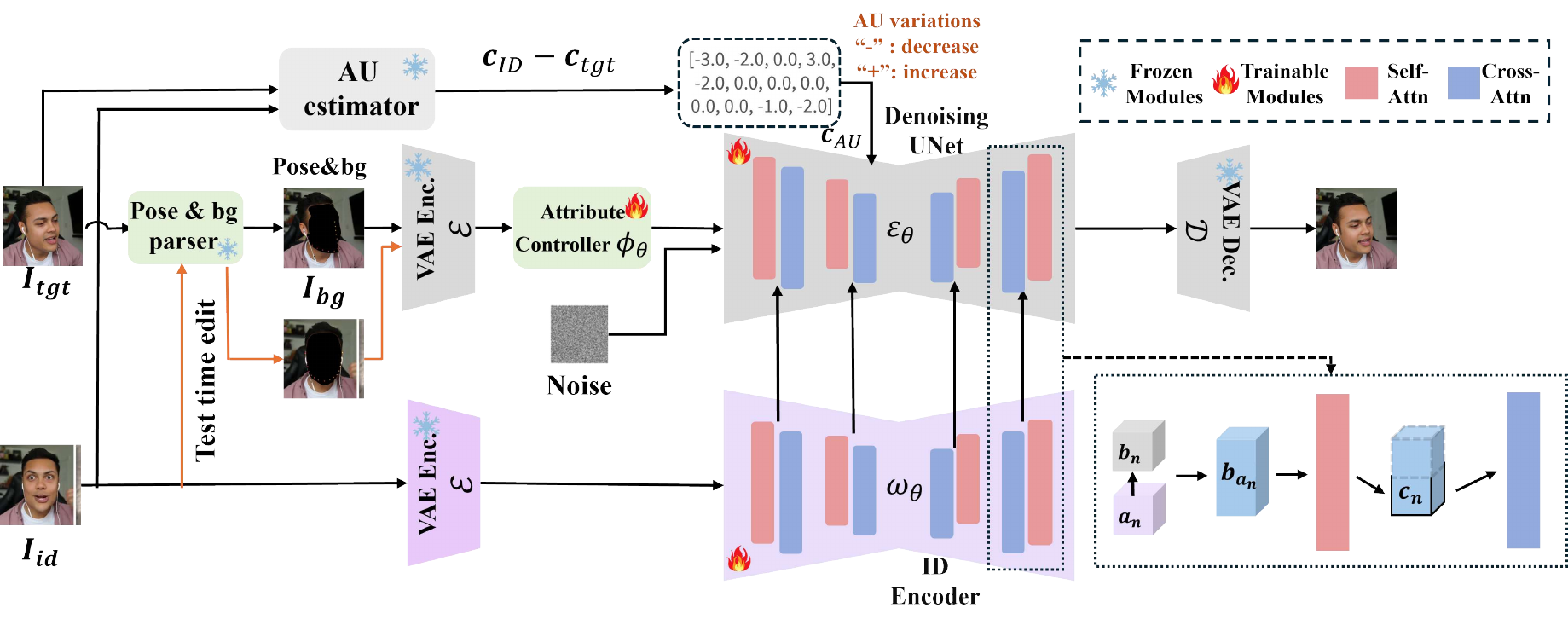}
    \caption{\textbf{Overview of MagicFace.} During training, a pair of images with the same identity but different pose, backgrounds, and expressions is used, respectively as the identity image and the target. AU variations are computed by an estimator and then sent into the denoising UNet as an AU condition. Pose and background of the target are parsed into an image condition independently, dealt with an Attribute Controller, and then input to the denoising UNet. ID encoder takes in the encoded identity image to edit for target AUs, where features in each transformer block are merged into the corresponding ones of the denoising UNet via self-attention. During inference, the conditional image will be parsed from the identity image.}
    \label{fig:framework}
\end{figure*}


\subsection{Preliminariy}

\noindent \textbf{Stable Diffusion.} Both the denoising UNet and ID encoder of our method inherit the same architecture as well as parameters of the denoising UNet in Stable Diffusion (SD). SD is developed from the latent diffusion model (LDM) \cite{rombach2022high}, which introduces learning of feature distributions to reduce computational complexity. It contains a fixed autoencoder to map input images into feature maps with a smaller size. Formally, for an image $\mathbf{x}$ given, the encoder maps it into a latent representation via $\mathbf{z}=\mathcal{E}(\mathbf{x})$ and the decoder reconstructs it by $\hat{\mathbf{x}}=\mathcal{D}(\mathbf{z})$. The denoising UNet in SD learns to denoise noise $\boldsymbol{\epsilon}$, which is normally distributed into $\mathbf{z}$. During training, the latent map $\mathbf{z}$ is added with Gaussian noise in $t$ timesteps and produces a noised latent $\mathbf{z}_t$. The UNet is trained to predict the noise added, optimized by the following objective:

\begin{equation}
\mathbf{L}=\mathbb{E}_{\mathbf{z}_t, c, \epsilon, t}\left(\left\|\epsilon-\epsilon_\theta\left(\mathbf{z}_t, c, t\right)\right\|_2^2\right),
\end{equation}

\noindent where $\epsilon$ denotes the denoising UNet and $c$ represents condition embeddings. In the original SD, $c$ is produced by a text encoder CLIP ViT-L/14 \cite{alexey2020image} to enable text-to-image generations. During inference, $\mathbf{z}_T$ is sampled from a normal distribution as the noised latent at timestep $T$, and then it is denoised iteratively into $\mathbf{z}_0$ by a deterministic sampling process like DDPM \cite{ho2020denoising}, DDIM \cite{song2020denoising}. Then $\mathbf{z}_0$ will be reconstructed by the decoder into a generated image. 

\noindent \textbf{AU variations VS absolute AUs.} Stepping back and reconsidering the conditioning inputs, using absolute AUs should be an intuitive choice. However, this condition configuration has a drawback: the model must estimate the actual AUs of the input image to decide whether to edit it. From an application standpoint, we are required to supply a value that must exactly match the corresponding AU in the source image, even if no changes are intended. AU variations, as opposed to absolute AUs, represent the intended change in specific action units. This aligns with the definition of action units, which indicates the activation state of facial muscles. Therefore, in our model, we choose the difference between target AUs $\mathbf{c}_{tgt}$ and source AUs $\mathbf{c}_{ID}$ as the input condition, denoted as:

\begin{equation}
\mathbf{c}_{AU} \triangleq \mathbf{c}_{ID}-\mathbf{c}_{tgt}
\end{equation}

Using AU variations as conditions offers several advantages. Firstly, it is intuitive and user-friendly. For instance, if we only wish to suppress AU10 (Upper Lip Raiser), we can assign any real negative value to this AU while setting all other values to zero. On the other hand, compared to the full set of target AUs, the values in AU variations are zero-centered, providing more meaningful information for guiding expression editing and stabilizing the training process. With AU variations, the model learns to edit and reconstruct facial regions based on both non-zero and zero values, reducing the complexity of preserving action units.

\subsection{Architecture}\label{sec:arc}

\textbf{Attribute Controller.} As shown in Fig. \ref{fig:framework}, Attribute Controller takes in the latent representation of an image where the face is masked and the face pose is outlined by landmarks. This design is to solve the ill-posed problem: in videos shot in wild settings, there are no image pairs with identical backgrounds and poses, while expressions undergo significant changes for the same person, which means no ground truth of the image with target AUs is available. However, we can find numerous images of the same person with different backgounds, pose, and facial expressions. As an alternative, we separate the background and pose as an independent condition of the identity by parsing out the background image, which will prevent the model from learning all the information through the input identity image. On the parsed background image, we draw the contour to highlight the head pose in case some backgrounds may be black. It is theoretically possible that the denoising UNet will learn background and pose from the identity. Empirically, we find the network learns to use the Attribute Controller for background and pose information and does not rely on the ID encoder. We hypothesize this is because the conditional image is pixel-aligned with the ground truth and thus more easily used by the model.

Attribute Controller is a convolution layer with 4 $\times$ 4 kernels, 2 $\times$ 2 strides, and 4 channels to align the latent of the conditional image with the same resolution as the noise input. The processed conditional image is then appended to the noisy latent together as the input to the denoising UNet.

\noindent\textbf{ID encoder.} We first evaluated the vanilla ControlNet \cite{zhang2023adding} for identity preservation. As illustrated in Fig. \ref{fig:abl}, we observed that ControlNet struggles to preserve appearance consistency when generating human images coherent to the AU prompt, making it unsuitable for our editing task. Many approaches employ the CLIP image encoder \cite{radford2021learning} to encode image conditions, but in our experiments, it doesn't adequately tackle problems associated with maintaining consistency in details. CLIP is designed to align semantic, high-level features with text; as a result, it will lose many details in the generated images. On the other hand, recent studies \cite{xu2024ootdiffusion,hu2024animate,cao2023masactrl} have highlighted the significant role of self-attention layers in diffusion models in maintaining the details of generated images. Building on these findings, we conducted an experiment on self-attention for identity control. In this setup, both the identity image and the noisy image are processed through the self-attention layers. A key observation from this architecture is that it inherently facilitates appearance resemblance between the two images. One possible explanation is that the self-attention layers in the UNet play a crucial role in spatially transmitting appearance information. As a result, they can function as a deformation module, enabling the generation of visually similar images with varying geometric structures.


Given the above observations, we design a feature extraction encoder $\omega_{\theta}$ for the identity image, which is aimed to preserve detailed face attributes, inspired by \cite{hu2024animate}. This encoder shares the same architecture as the denoising UNet $\phi_{\theta}$. Specifically, let $\mathbf{s}=\mathcal{E}(\boldsymbol{I}_{id})$ denote the encoded latent feature sent to the ID encoder $\omega_{\theta}$ where $\boldsymbol{I}_{id}$ is the identity image. The features extracted from $\omega_{\theta}$ will be merged into the denoising UNet $\epsilon_{\theta}$ through self-attention. Along with AU conditions, the ID encoder and the denoising UNet are jointly optimized by:

\begin{equation}
\label{equ:loss}
\mathcal{L}_{AUEdit}=\mathbb{E}_{\mathbf{z}_t, \mathbf{s}, \mathbf{c}_{AU}, \mathbf{g}, \epsilon, t}[\|\epsilon-\epsilon_\theta(\mathbf{z}_t, t, \omega_{\theta}(\mathbf{s}), \mathbf{c}_{AU}, \phi_{\theta}(\mathbf{g}))\|_2^2],
\end{equation}

\noindent where $\mathbf{g}=\mathcal{E}(\boldsymbol{I}_{bg})$ denotes the encoded latent feature sent to the Attribute Controller $\phi_{\theta}$, $\boldsymbol{I}_{bg}$ is the background image and $\mathbf{c}_{AU}$ represents the encoded AU condition. We omit the AU encoder in the framework as it's simply a linear layer to map the input AU into a feature with the same dimension as the time conditions in SD (details are deferred to Sec. \ref{sec:abla}). The AU condition is added to the time-embedding condition in $\phi_{\theta}$. To merge identity features from $\omega_{\theta}$ into $\phi_{\theta}$ with self-attention, we first dive into the transformer block \cite{vaswani2017attention} of the ID encoder and denoising UNet, identifying each pair of feature maps that serve as inputs for the respective self-attention layers. For ease of explanation, we temporarily use $\mathbf{a}_n \in \mathbb{R}^{h_n \times w_n \times c_n}$ and $\mathbf{b}_n \in \mathbb{R}^{h_n \times w_n \times c_n}$ to denote the $n$th pair of feature maps from ID encoder and denoising UNet. In the denoising UNet, in each block the feature maps from both sides are first concatenated by $\mathbf{b}_{\mathbf{a}_n}=\mathbf{a}_n \bigoplus \mathbf{b}_n \in \mathbb{R}^{h_n \times 2w_n \times c_n}$, then we replace $\mathbf{b}_n$ with $\mathbf{b}_{\mathbf{a}_n}$ as the input to the self-attention layer. After coming out from the self-attention layer, we crop out the first half of the output feature map into $\mathbf{c}_n$ of resolution $h_n \times w_n \times c_n$ before input to the subsequent cross-attention layer.

\subsection{AU dropout}

To improve the controllability of our method, we incorporate an AU dropout operation during training, enabling classifier-free guidance \cite{ho2021classifier} for AU conditions. This strategy has been widely applied in conditional image generation to balance the trade-off between quality and diversity in images generated by latent diffusion models. In our training process, we randomly drop the AU as all zeros $\mathbf{c}_{uncon\_AU} = \varnothing$. To distinguish the dropped AUs from vectors representing no AU changes, we add a small amount of Gaussian noise to the vectors representing no AU changes, denoted as $\mathbf{c}_{zero\_AU}'= \mathbf{c}_{zero\_AU} + \epsilon, \epsilon \sim \mathcal{N}(\mu, \sigma^2)$. We empirically set $\sigma=0.2$. At the inference stage, we use a guidance scale $\alpha>1$ to modify the intensity of conditional control on the predicted noise:

\begin{equation}
\label{equ:guidance}
\hat{\epsilon}_\theta\left(\mathbf{z}_t, \mathbf{c}_{AU}\right)=\epsilon_\theta\left(\mathbf{z}_t, \varnothing\right) + \alpha \cdot\left(\epsilon_\theta\left(\mathbf{z}_t, \mathbf{c}_{AU}\right)-\epsilon_\theta\left(\mathbf{z}_t, \varnothing\right)\right).
\end{equation}

\noindent Note that we omit other conditions including $t, \omega_{\theta}(\mathbf{s}), \phi_{\theta}(\mathbf{g})$ compared with Eq. \ref{equ:loss} to explain more intuitively.

We empirically configure the AU dropout ratio to 10\% during training, i.e., 10\% AUs are set to $\varnothing$. Based on our ablation study (see Sec. 4.3), the optimal value of the guidance scale $\alpha$ is typically in the range of 1.5 to 3.5.

\subsection{Training Strategy}

As introduced in Section \ref{sec:arc}, the background and pose are formed into an independent image condition, so in the training stage, we use the background of the target image parsed by \cite{yu2021bisenet}. We draw the pose contour on the background image by highlighting the 14 landmarks \cite{bulat2017far} around the chin of the face. Then, in the inference time for editing, we use the background and pose parsed from the identity image to enable the generated results to be consistent with the identity in terms of background. We initialize the denoising UNet and the ID encoder with the pre-trained weights from SD. Both the weights of the denoising UNet and ID encoder are updated independently during training. Note that the ID encoder is only an encoder without time conditions, which needs only one step forward process before the multiple denoising steps in inference.

\section{Experiments}

\subsection{Implementations}

\textbf{Experimental settings.} We collect 30K image pairs from the Aff-Wild dataset \cite{kollias2019deep} by decomposing the videos into frames to train our model. Each pair has the same identity but different expressions, wherein at least one of the poses or backgrounds is highly likely to be different by ensuring that two sampled pairs maintain a certain distance in the video. We use LibreFace \cite{chang2024libreface} to compute 12 AUs variation of the pair images and use Stable Diffusion 1-5 base \footnote{https://huggingface.co/stable-diffusion-v1-5/stable-diffusion-v1-5} to initialize the weights. Experiments are conducted on 4 NVIDIA A100 GPUs. Training is conducted for 100,000 steps with a batch size of 2 on each GPU. The learning rate is set to 1e-5.

\noindent\textbf{Evaluation Criteria.}
We assess the performance of MagicFace by examining the edited image from four perspectives: the accuracy of AU intensity, identity preservation, background preservation, and head pose preservation. We measure the AU values of the edited image $Y_{est}$ and compare them with the intended target values $Y_{tar}$ by computing the mean squared error (MSE). LibreFace \cite{chang2024libreface} is used to estimate AU intensities of the generated images. To assess the similarity of identity, we measure the distance of embeddings between the edited image $\boldsymbol{I}_{tar}$ and the identity image $\boldsymbol{I}_{id}$, where the embedding is extracted from a pre-trained ID recognition model \footnote{https://github.com/ageitgey/face\_recognition}. For background preservation, we employ the root pixel-wise mean squared error (RMSE) between the edited image and the identity image, where the face area of both is masked to eliminate the interference of face expression difference. For head pose preservation, we compute head pose coefficients of the FLAME model \cite{li2017learning} and compute their root mean squared error (RMSE).

\begin{figure*}[htbp]
    \centering
    \includegraphics[width=1.0\textwidth]{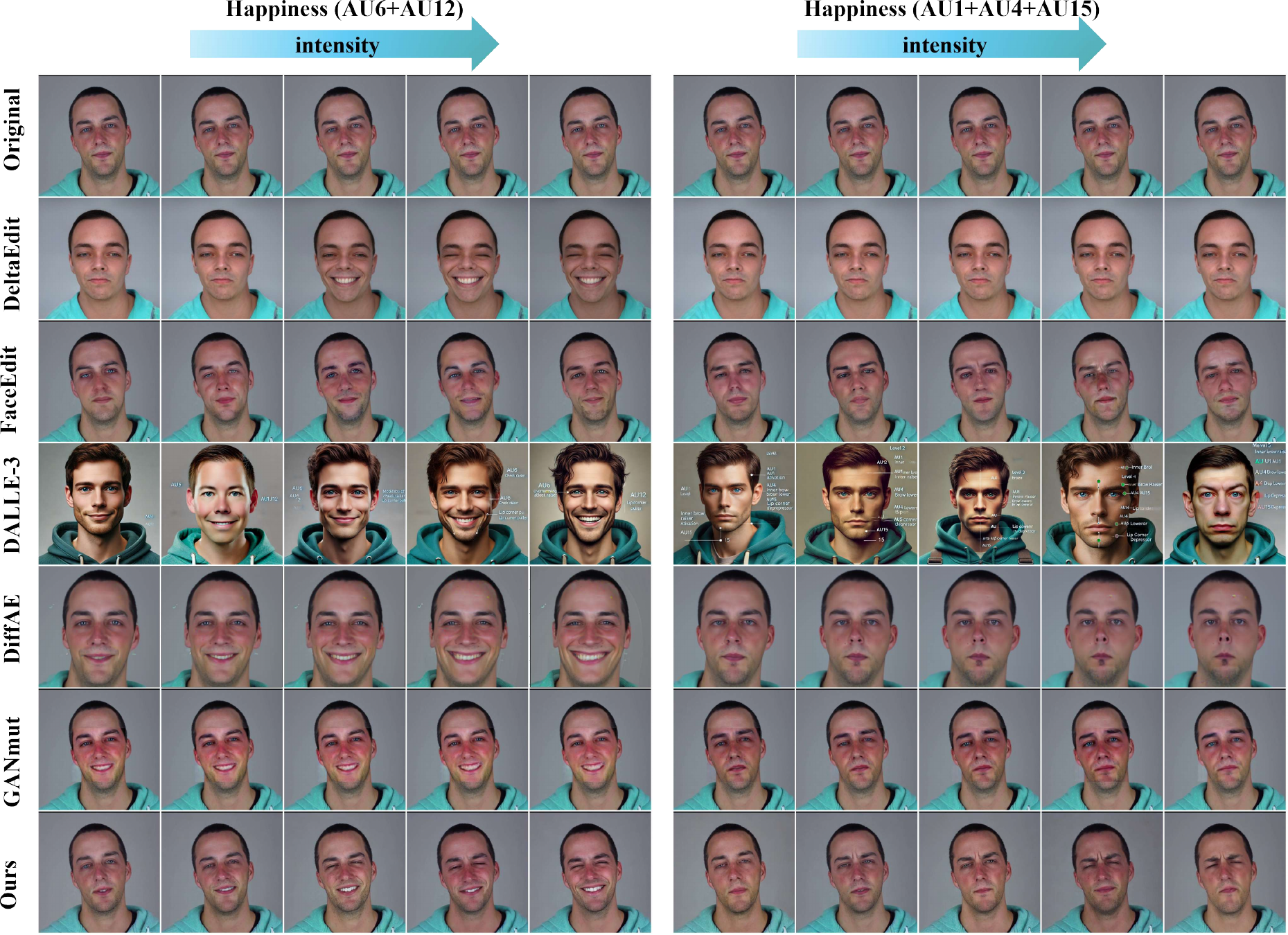}
    \caption{\textbf{Qualitative comparison for continuous facial expression editing.} Our method excels in maintaining exceptional detail features of the face, while allowing flexible, fine-grained control over the expression intensity. Please zoom in for a more detailed observation.}
    \label{fig:results}
\end{figure*}

\subsection{Results}

\noindent\textbf{Continuous Facial Expression Editing.} We edit 20 identities with variations on individual AUs and on AU combinations. We use 5 intensity variation levels for each single AU out of 12 AUs, and for AU combinations, we use an in total of 387 combinations. This leads to a total number of $20 \times (12 \times 5 + 387)= 8940$ edited images. Then we compare our method with several representative approaches related to facial expression editing, including DeltaEdit \cite{lyu2023deltaedit}, FaceEdit\footnote{https://github.com/ototadana/sd-face-editor?tab=readme-ov-file}, DALLE-3\footnote{https://openai.com/index/dall-e-3/}, DiffAE \cite{preechakul2021diffusion}, and GANmut \cite{d2021ganmut}. For qualitative comparison, we show results on two emotions, i.e., happiness and sadness. For DeltaEdit and FaceEdit, which do not support AU descriptions, we use text prompts that contain emotional keywords together with adjectives describing the strength of the expression, for example, mild happiness and intense happiness. DALLE-3 can understand AU instructions, so we design prompts such as \textit{add intensity level 2 (level range [1, 10]) of AU1, AU4, and AU15 to this face} in order to produce results that align with AU intensity as much as possible. DiffAE provides a one-dimensional control for smiling. Its positive direction is used to represent happiness, and its negative direction is used to represent sadness. GANmut is conditioned on Valence and Arousal. We follow the emotion-to-VA mapping provided in their work to generate the target expressions. In our method, we adopt the AUs associated with the two emotions following \cite{ekman1978facial}. Specifically, AU6+AU12 is for happiness, and AU1+AU4+AU15 is for sadness. Fig.~\ref{fig:results} presents the comparisons for the happiness and sadness expressions, and TABLE~\ref{tab:comp} summarizes the quantitative results for all editing settings.

Our model surpasses the majority of existing methods in overall generation quality. Approaches like DeltaEdit and DELLA-3 are unable to effectively preserve the identity of the input image. DELLA-3 performs well in continuously increasing the expression intensity, and it can also interpret AU semantic meanings, but it can only generate the image with a similar style to the input image, where the identity, background, and pose are likely to be changed. Some methods, including DiffAE, GANmut, DeltaFace, are unable to continuously control the intensity of expression specifically for sadness. The results generated by FaceEdit are the most similar to ours, but they fail to preserve facial features well enough. Moreover, FaceEdit is unable to support the editing of other customized facial expressions. TABLE \ref{tab:comp} provides a quantitative demonstration of the comparison, where we can observe that DELLA-3 and our model obtain much lower error in terms of the AU intensity accuracy. FaceEdit and GANmut show better scores in terms of identity preservation as well as image similarity, corresponding to the observation from Fig. \ref{fig:results}. Overall, our method provides strong capability and flexibility in controlling the intensity and continuity of facial expressions.

\begin{table}[h!]
\centering
\caption{\textbf{Comparison of methods in terms of AU accuracy, identity preservation, and image similarity.} N/A denotes incomputable. The best and second best results are reported in \textbf{bold} and [square brackets], respectively.} 
\label{tab:comp}
\begin{tabular}{lcccc}
\toprule
\multirow{2}{*}{Method} & AU   & ID  & \makecell[c]{back- \\ground}  & \makecell[c]{head \\pose} \\
                          \cline{2-5} 
                       ~       & MSE ($\downarrow$)   & L2 ($\downarrow$)  &RMSE ($\downarrow$)  & RMSEs ($\downarrow$) \\
\midrule
DeltaEdit \cite{lyu2023deltaedit}  & N/A      &0.743      &0.126    &0.075 \\  
FaceEdit   & N/A      &0.513       &\textbf{0.044}      &\textbf{0.032} \\
DALLE-3    & 0.512      &0.763       &0.224    &0.208 \\
DiffAE \cite{preechakul2021diffusion}    & N/A       & 0.579  &0.161     &0.183\\
GANmut \cite{d2021ganmut}   & N/A      &[0.495]       &0.059    &\textbf{0.032} \\
\hline
\textbf{MagicFace (Ours)} &\textbf{0.261}       &\textbf{0.473}       &\textbf{0.044}     &\textbf{0.032} \\ 

\bottomrule
\end{tabular}
\end{table}

\begin{table}[h!]
\centering
\caption{\textbf{Quantitative comparison of discrete expression editing performance.} The best and second best results are reported in bold and [square brackets], respectively.} 
\label{tab:discrete}
\begin{tabular}{lcccc}
\toprule

Method & \multicolumn{1}{c}{Emotion} & ID & background & head pose \\

       & Acc (\%) $\uparrow$ & L2 ↓ & RMSE ↓ & RMSE ↓ \\
\midrule
GANmut \cite{lyu2023deltaedit}  & [90]      &0.497      &0.057    &\textbf{0.035} \\  
EmoStyle \cite{azari2024emostyle}   & 63      &0.549       &0.071      &0.087 \\
InstantDrag \cite{shin2024instantdrag}   & 66      &[0.474]       &0.051    &0.069 \\
DragDiffusion \cite{shi2024dragdiffusion}    & 27       & 0.480  &\textbf{0.048}     &0.038\\
SDE-Drag \cite{d2021ganmut}   & 46      &0.495       &[0.050]    &[0.037] \\
Step1X-Edit  \cite{liu2025step1x}   & 80     & 0.613  &\textbf{0.048}     &0.038  \\
\hline
\textbf{MagicFace (Ours)} &\textbf{92}    &\textbf{0.466}       &\textbf{0.048}     &[0.037] \\ 

\bottomrule
\end{tabular}
\end{table}

\begin{figure}[htbp]
    \centering
    \includegraphics[width=0.48\textwidth]{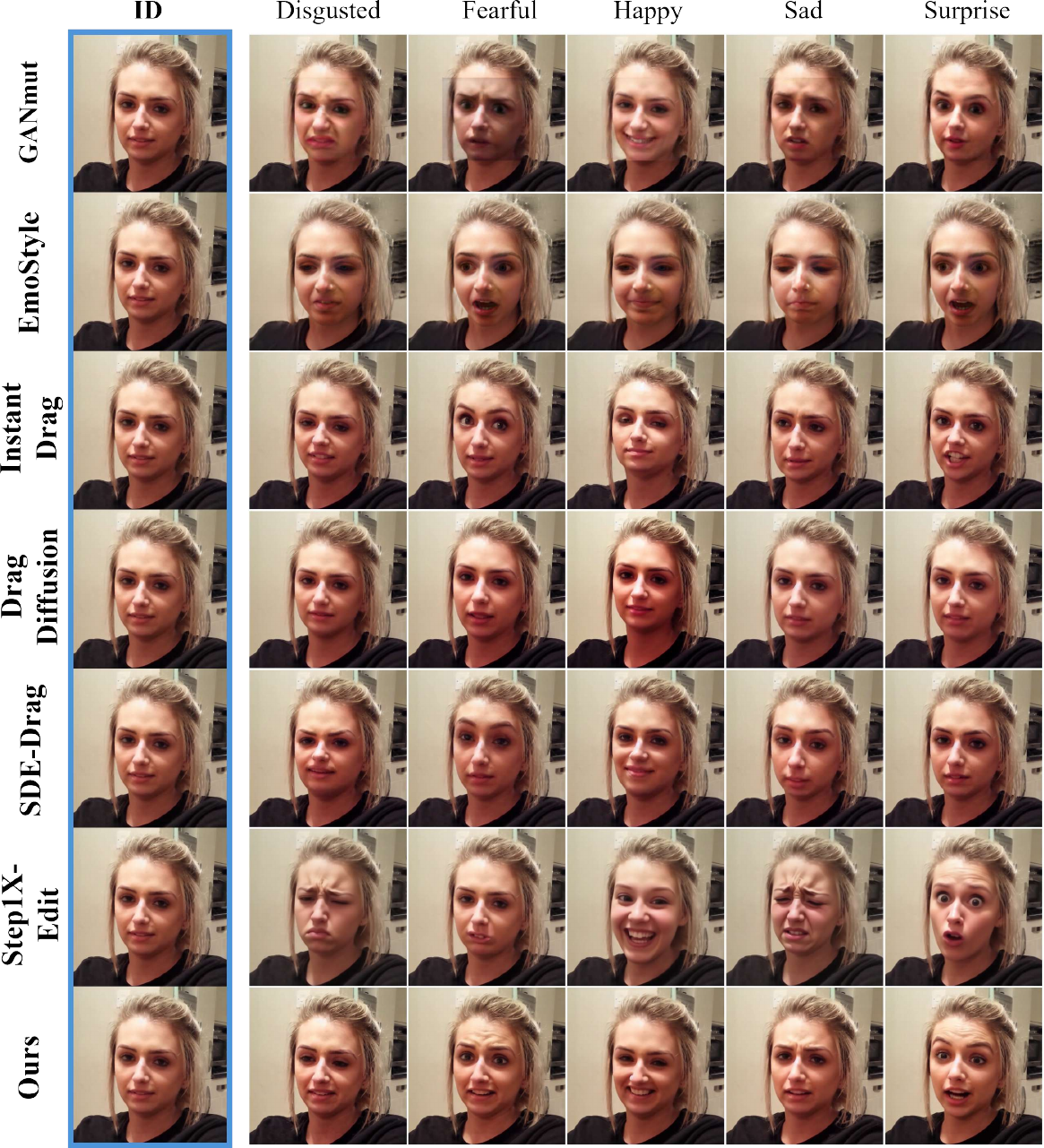}
    \caption{\textbf{Qualitative comparison with representative methods on discrete facial expression editing.} The leftmost column shows the input images used as the editing source, and the remaining columns display the edited results for each target expression.}
    \label{fig:discrete}
\end{figure}

\begin{figure*}[htbp]
    \centering
    \includegraphics[width=0.99\textwidth]{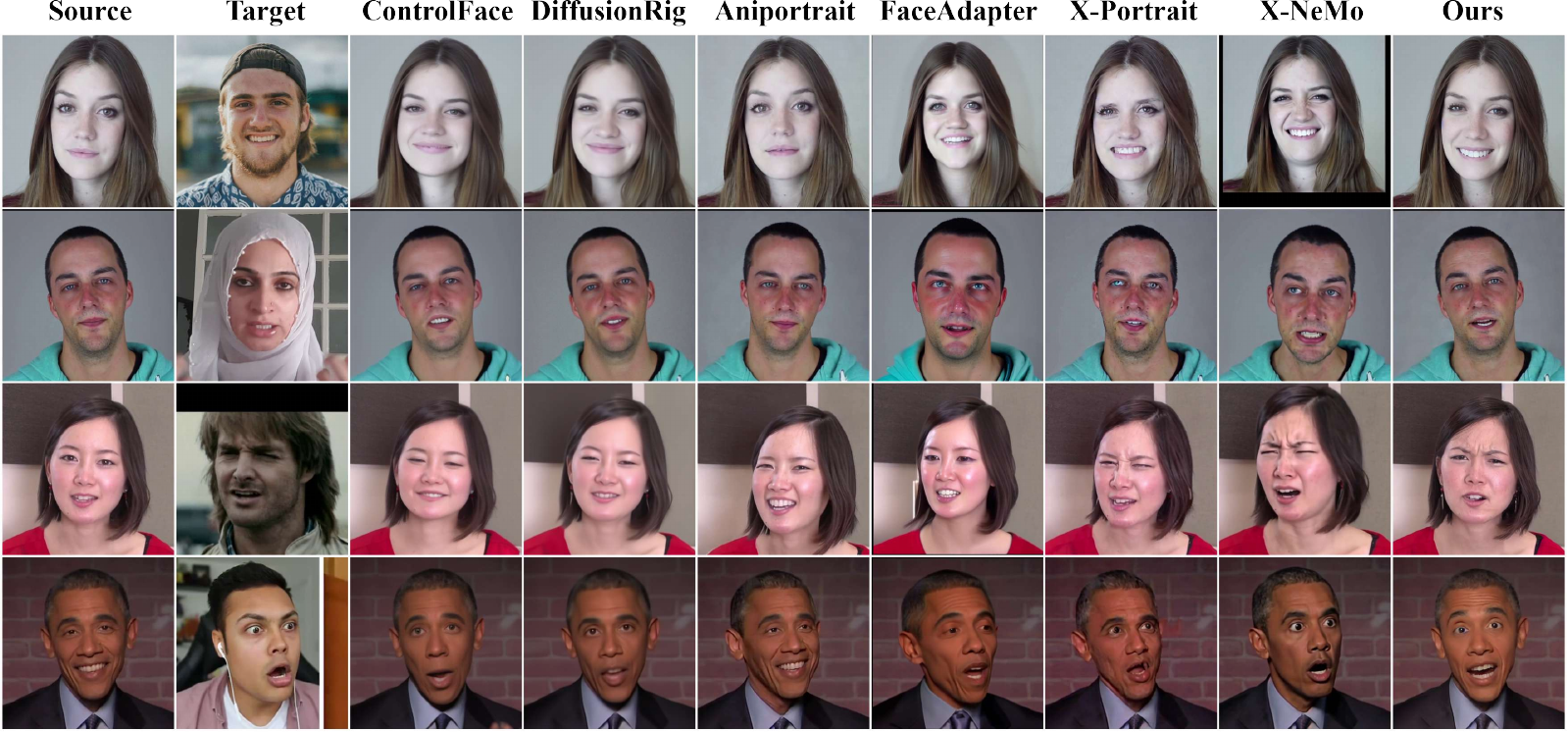}
    \caption{\textbf{Qualitative comparison of facial expression transfer.} The first column shows the source identities, and the second column displays the target expressions. AniPortrait, FaceAdapter, X-Portrait, and X-NeMo are face-reenactment methods, so their outputs inherit both the target expression and its pose.}
    \label{fig:transfer}
\end{figure*}

\noindent \textbf{Discrete Facial Expression Editing.} We further evaluate our method on discrete facial expression editing. To make the comparison more comprehensive, we extend the set of discrete expressions beyond happiness and sadness and include three additional emotions, namely fear, surprise and disgust. We compare our model with several representative methods across different editing paradigms: activation–arousal based methods (GANmut \cite{d2021ganmut}, EmoStyle \cite{azari2024emostyle}), dragging-based editing methods (DragDiffusion \cite{shi2024dragdiffusion}, InstantDrag \cite{shin2024instantdrag}, SDE-Drag \cite{nie2023blessing}), and a recent text-driven method Step1X-Edit \cite{liu2025step1x}. We randomly sample 20 test images and evaluate the editing performance from four perspectives: emotion accuracy, identity similarity, background consistency and head pose consistency. Emotion accuracy is estimated using the pretrained EmoNet classifier \cite{toisoul2021estimation}, while the other three metrics follow the evaluation procedures described in Sec. IV-A.

The qualitative and quantitative results are reported in Fig. \ref{fig:discrete} and TABLE \ref{tab:discrete}. As shown, our method consistently achieves the highest emotion accuracy across all discrete expressions, demonstrating its more precise control over the intended emotional changes. Furthermore, our approach preserves facial identity noticeably better than competing methods. Text-driven approaches such as Step1X-Edit \cite{liu2025step1x} tend to unintentionally alter identity, while dragging-based methods may introduce pose inconsistencies. InstantDrag is particularly prone to such artifacts as it does not apply any masking mechanism. For example, when editing expressions such as happy, it may interpret the dragging direction as a head movement, leading to undesirable changes in pose (see the third row in Fig. \ref{fig:discrete}). In addition, our method produces more realistic and fine-grained expression details. As illustrated in Fig. \ref{fig:discrete}, especially in the fear and surprise cases, our results exhibit subtle cues such as forehead wrinkles that commonly appear in natural expressions. These details are not observed in GANmut \cite{lyu2023deltaedit}, even though it can generate expressions with relatively strong intensity.

\begin{table}[h!]
\centering
\caption{\textbf{Quantitative comparison of facial expression transfer performance.} The best and second best results are reported in bold and [square brackets], respectively.} 
\label{tab:transfer}
\begin{tabular}{lccc}
\toprule

Method & \multicolumn{1}{c}{Expression} & ID & AU \\
       & EXP-SIM $\uparrow$ & L2 ↓ & MSE ↓  \\
\midrule
ControlFace \cite{jang2025controlface}  & 0.47      &0.488      &0.482     \\  
DiffusionRig \cite{ding2023diffusionrig}   & 0.45      &0.485       &0.502       \\
AniPortrait \cite{wei2024aniportrait}   & 0.41      &[0.477]       &0.535     \\
FaceAdapter \cite{han2024face}    & 0.52       & 0.565  &0.431     \\
X-Portrait \cite{xie2024x}   & 0.54      &0.524       &0.392     \\
X-NeMo \cite{zhao2025x}   & \textbf{0.66}     & 0.513  &\textbf{0.253}       \\
\hline
\textbf{MagicFace (Ours)} &[0.61]    &\textbf{0.471}       &[0.288] \\ 

\bottomrule
\end{tabular}
\end{table}

\noindent \textbf{Facial Expression Transfer.} Our method is further evaluated on a facial expression transfer task. Since facial expressions can be manipulated through various types of conditioning, several representative approaches across different paradigms are included for comparison. For 3D-based control, ControlFace \cite{jang2025controlface} and DiffusionRig \cite{ding2023diffusionrig} are selected. For landmark-based control, AniPortrait \cite{wei2024aniportrait} and FaceAdapter \cite{han2024face} are considered. For drive-image-based control, X-Portrait \cite{xie2024x} and X-NeMo \cite{zhao2025x} are evaluated. AniPortrait, FaceAdapter, X-Portrait, and X-NeMo are face reenactment methods in which expression and head pose are inherently entangled. Therefore, pose and background consistency are not evaluated in this setting. Instead, the comparison focuses on three aspects: the precision of expression transfer, identity similarity, and AU accuracy. The accuracy of expression transfer is measured using the pretrained emotion encoder EmoNet \cite{toisoul2021estimation}. Specifically, the mean concordance correlation coefficient and Pearson correlation coefficient of valence and arousal are computed, and an overall emotion similarity score (EXP-SIM) is obtained. This metric reflects how precisely fine-grained expression variations are captured.

The qualitative and quantitative comparisons are presented in Fig.~\ref{fig:transfer} and TABLE~\ref{tab:transfer}. Our method achieves the strongest identity preservation and obtains competitive expression transfer accuracy. Although X-NeMo reaches higher EXP-SIM values, visual inspection reveals several limitations. As shown in the fourth row of Fig.~\ref{fig:transfer}, X-NeMo may inherit identity traits from the driving image, resulting in noticeable identity leakage, and tends to distort the head size of the source subject (other rows). In contrast, our method consistently maintains the identity characteristics of the source while transferring the target expression. Moreover, compared with 3D-based and landmark-based methods, our approach better captures subtle and fine-grained expression cues. For instance, in rows three and four of Fig.~\ref{fig:transfer}, expressions involving frowning or widening of the eyes naturally produce forehead wrinkles. These details are faithfully reproduced by our method but are missing in the 3D-based and landmark-based approaches, which have limited representational capacity for such nuances. This highlights the advantage of AU-based conditioning for achieving both precise and expressive facial animation.

\begin{figure}[htbp]
    \centering
    \includegraphics[width=0.48\textwidth]{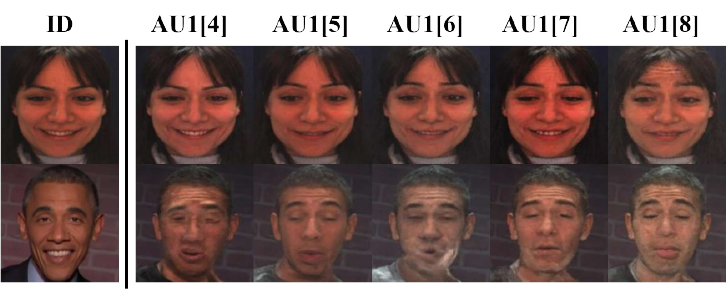}
    \caption{\textbf{Results of using the laboratory dataset for training.} In a lab setting, the model cannot generalize to images from natural settings (the second row).}
    \label{fig:disfa}
\end{figure}

\begin{figure*}[htbp]
    \centering
    \includegraphics[width=1.0\textwidth]{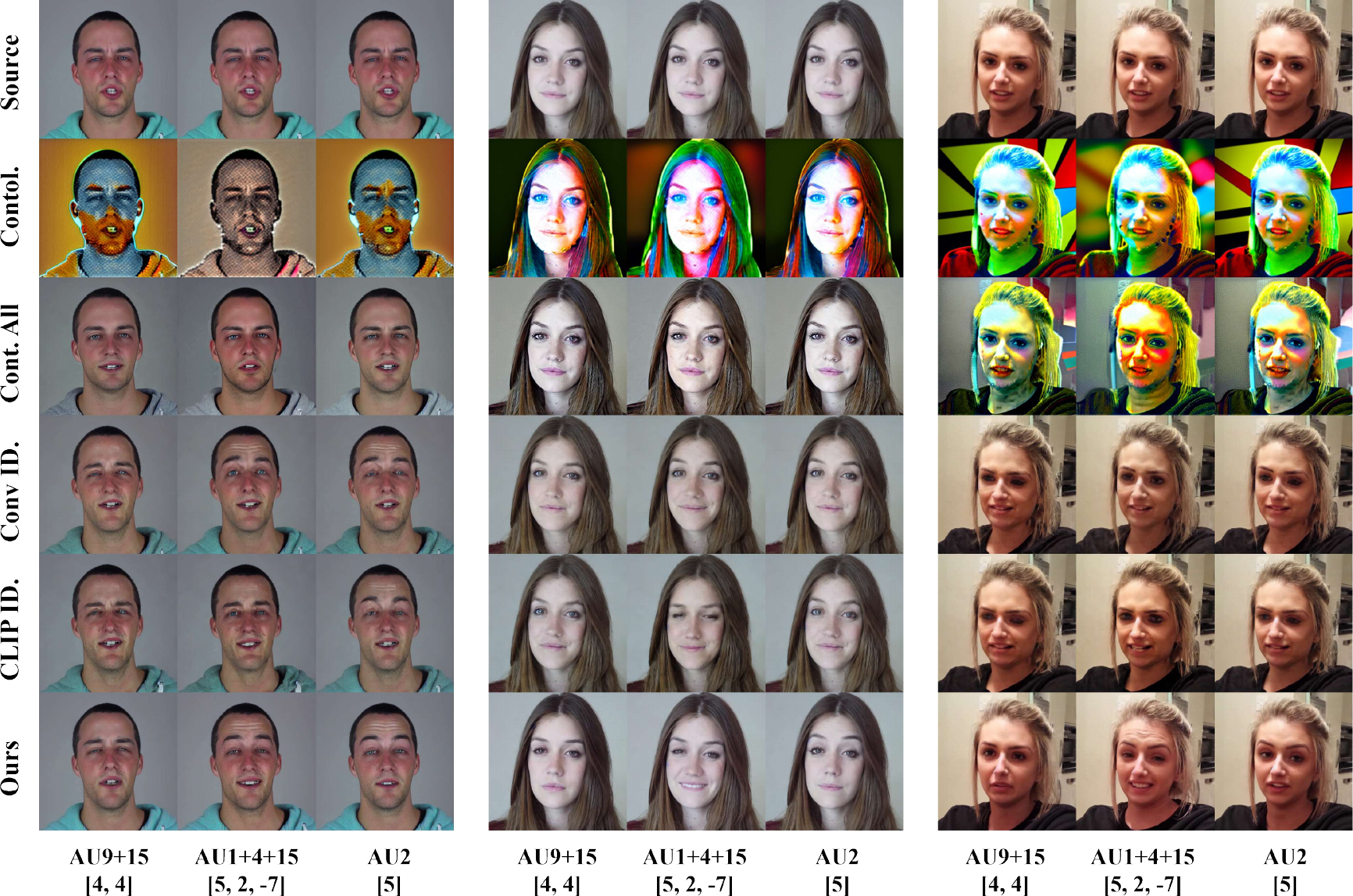}
    \caption{\textbf{Ablation study for different model architectures.} Zoom in to view facial details.}
    \label{fig:abl}
\end{figure*}

\subsection{Ablation Study} \label{sec:abla}
\textbf{Training data.} We investigate the effectiveness of using in-the-wild data to train our model by comparing with the model trained on DISFA \cite{mavadati2013disfa}. DISFA provides AU annotation manually annotated by trained people, so literally using this dataset to train could produce more accurate AU conditions than using AU conditions from AU estimation tools. However, as shown in Fig. \ref{fig:disfa}, training on lab data can only generalize to new data in laboratory settings and cannot generalize to data from natural settings. The reason for this can be attributed to the overly simplistic data patterns: simple background, only frontal face, and a limited number of individuals in DISFA. Although the AU estimator may bring inaccuracies in the estimated AU intensity of the training data, we can observe in Fig. \ref{fig:demo} and \ref{fig:results} that it can still generate edited images with high coherence to the AU prompt.

\begin{table}[h!]
\centering
\caption{\textbf{Ablation study on model architecture.} The best and second best results are reported in \textbf{bold} and [square brackets], respectively.} 
\label{tab:abl2}
\begin{tabular}{lcccc}
\toprule
\multirow{2}{*}{} & AU    & ID   &\makecell[c]{back- \\ground}  &\makecell[c]{head \\ pose} \\
                          \cline{2-5} 
                       ~       & MSE ($\downarrow$)    & L2 ($\downarrow$)  & RMSE ($\downarrow$)  & RMSE ($\downarrow$) \\
\midrule
ControlNet  & 0.725      &0.664         &0.164      &\textbf{0.032} \\  
Cont. All   & 0.673       &0.610         &0.130      &[0.033] \\
Conv ID.  & 0.462        &0.574         &[0.070]       &[0.033] \\  
CLIP ID.   & [0.406]     &[0.543]      &0.073     &\textbf{0.032} \\
Ours &\textbf{0.261}     &\textbf{0.473}    &\textbf{0.044}   &\textbf{0.032}\\ 

\bottomrule
\end{tabular}
\end{table}

\begin{table}[h!]
\centering
\caption{\textbf{Ablation study on AU encoders.} The best and second best results are reported in \textbf{bold} and [square brackets], respectively.} 
\label{tab:ablation}
\begin{tabular}{lcccc}
\toprule
\multirow{2}{*}{} & AU    & ID  & \makecell[c]{back- \\ground}  & \makecell[c]{head \\pose}\\
                          \cline{2-5} 
                       ~       & MSE ($\downarrow$)    & L2 ($\downarrow$)  &RMSE ($\downarrow$)  & RMSE ($\downarrow$) \\
\midrule
MLP+Conv  & \textbf{0.204}      &0.501       &[0.044]     &\textbf{0.032} \\  
ZeroAppend+Time   & 0.282      &[0.495]       &0.054     &\textbf{0.032}\\
Linear+Time (Ours) &[0.261]       &\textbf{0.473}       &\textbf{0.044}     &\textbf{0.032}\\ 

\bottomrule
\end{tabular}
\end{table}

\noindent \textbf{Model design.} To demonstrate the effectiveness of our model's design, we explore four alternatives, 1) Replacing the whole architecture with ControlNet \cite{zhang2023adding} (\textit{Control.}) and add our Attribute Controller to enable background embedding; 2) Copy the architecture of 1) but enable denoising trainable (\textit{Cont. All}); 3) Replacing the ID Encoder by a Conv layer (\textit{Conv ID.}); 4) Replacing the ID Encoder by a CLIP image encoder \cite{radford2021learning} (\textit{CLIP ID.}).  Results are shown in Fig. \ref{fig:abl} and TABLE \ref{tab:abl2}. Our design achieves the optimal performance. Design 1) and 2) show that ControlNet can not adapt to AU conditions, and it significantly alters the background and the appearance of the face. Design 2) and 4) show that using CLIP or Conv to extract features can preserve image similarity, but fail to preserve facial details. This further prevents them from precisely editing facial expressions based on AUs, as some facial details are lost. The edited results of these two experiments also differ from the identity image in some color details, like the hoodie and face color of the leftmost identity. We can also observe that in all of these ablations, head pose is well-preserved, which indicates the effectiveness of separating background and pose as a pixel-aligned condition to make the model focus on the face area.

\begin{figure*}[htbp]
    \centering
    \includegraphics[width=1\textwidth]{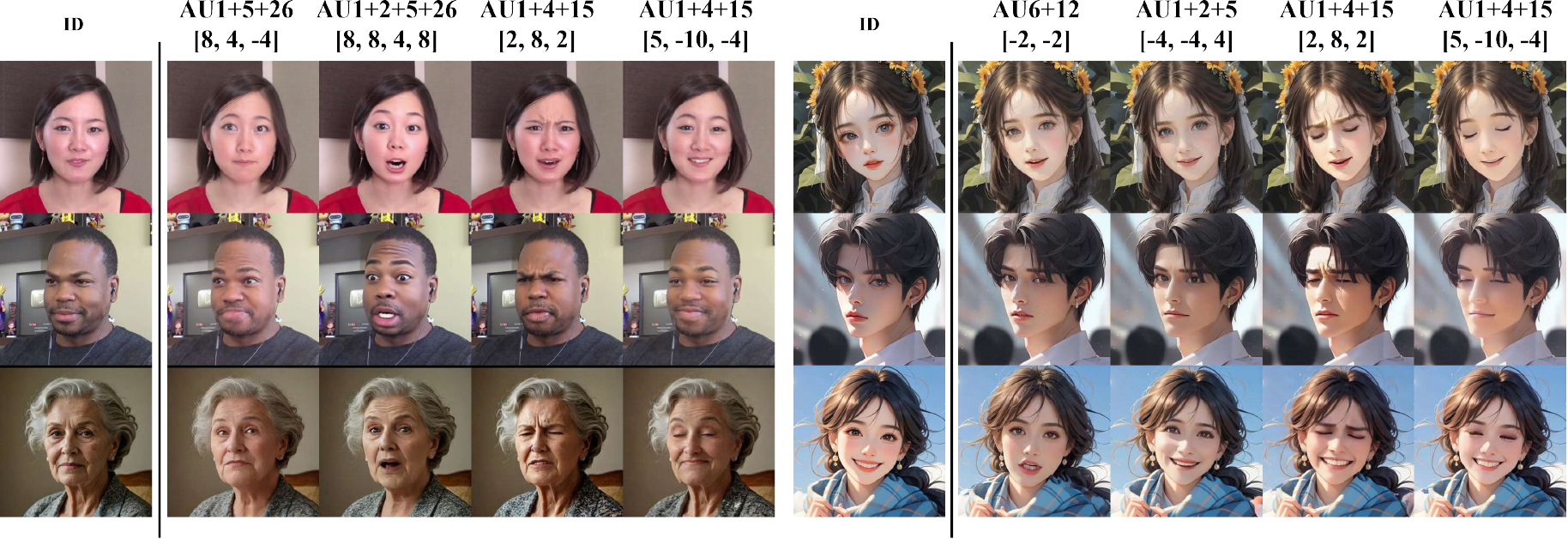}
    \caption{\textbf{Demonstration of out-of-domain testing for extreme expressions as well as unseen styles.} Some AUs are out of the range [0, 5]. The left column displays the editing results of real person photos with extreme expressions, and the right column displays those of the cartoon characters.}
    \label{fig:extreme}
\end{figure*}

\noindent \textbf{Out-of-domain editing.} It is worth noting that MagicFace demonstrates remarkable generalization to out-of-domain identity images with unseen AUs and characters, achieving impressive appearance controllability even without additional fine-tuning on the target domain. In the training stage, the utmost AU variation given a neutral expression is $\pm 5$ (AU annotation only in range [0, 5]), but our model can edit the range of AUs, almost extending from -10 to 10. This indicates that MagicFace has a strong capability to generate and explore contradictory and extreme expressions, as we can set AUs associated with contradictory emotional states to have opposite intensity variations and large values. On the other hand, on zero-shot results of applying our model to cartoon characters whose visual style is distinct from the training data of the real-human, our model can still ensure AU coherence with high-quality generation. We visualize the qualitative results in Fig. \ref{fig:extreme}.

\noindent \textbf{AU Encoder.} We investigate the influence of the AU encoder by conducting two additional experiments: 1) Apply an MLP to encode AU vectors and concatenate it to the first Conv layer of the denoising UNet (MLP+Conv). 2) Append zero values to the end of the AU prompt to make the input vector of the same size as the time embedding and add it to the time embedding (ZeroAppend+Time). Quantitative results are shown in TABLE \ref{tab:ablation}. As can be observed, the encoding method of AUs does not significantly affect the performance of the approach, so we choose an easier way by just applying a linear layer to the AU prompt and then adding it to the time embedding (Linear+Time).

\begin{figure*}[htbp]
    \centering
    \includegraphics[width=1\textwidth]{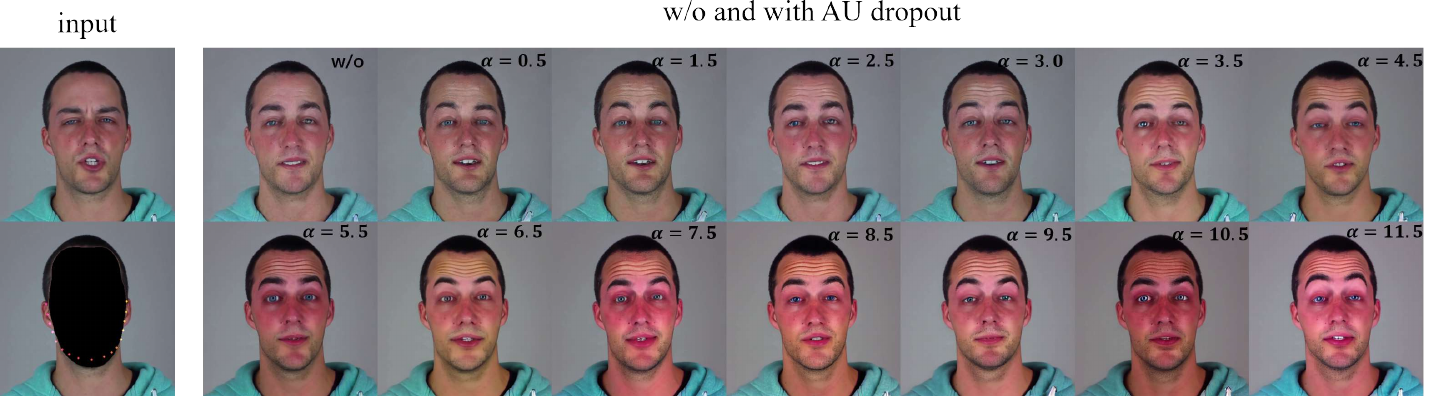}
    \caption{\textbf{Qualitative comparison of images edited by MagicFace trained without/with AU dropout and using different values of the guidance scale $\alpha$.} The AU variation to edit is AU4[-6]. Please zoom in for more details.}
    \label{fig:guidance}
\end{figure*}

\noindent\textbf{AU dropout.} We investigate the effects of AU dropout as well as the different values of the guidance scale $\alpha$. We first train our model without/with AU dropout. Then, for the model trained with AU dropout, we set $\alpha=0.5, 1.5, 2.5, 3.0, 3.5, 4.5, 5.5, 6.5, 7.5, 8.5, 9.5, 10.5, 11.5$, $12.5, 13.5$ for classifier-free guidance. During inference, we ensure that all other parameters, including the random seed, remain consistent to allow for a fair comparison. As illustrated in Fig. \ref{fig:guidance}, without AU dropout, classifier-free guidance is not supported, resulting in the worst generation quality where the target AU4[-6] is not obvious. For the model trained with dropout, when $\alpha > 1.0$, the opposite variation of AU4 related area (brow lower towards brow higher) becomes progressively more distinct as $\alpha$ increases. However, color distortion will appear when $\alpha \ge 4.5$ and becomes increasingly noticeable starting from $\alpha = 5.0$, as seen in the hoodie color, more evident beard, and more red face. TABLE \ref{tab:guidance} provides quantitative evidence supporting the effectiveness of our AU dropout. It also indicates that the optimal guidance scale is typically in the range of 1.5 to 3.5 in most cases. According to this study, we empirically set $\alpha=3.0$ for all test experiments.


\begin{table*}[!htbp]
\centering
\caption{\textbf{Ablation study of AU dropout and different guidance scale values.} The best and second best results are reported in \textbf{bold} and [square brackets], respectively. "BG" is the abbreviation for "background", and "HP" is the abbreviation for "head pose." The best results are reported in \textbf{bold}.} 
\label{tab:guidance}
\centering
\begin{tabular}{cc|cccc|cc|ccccccc}
\toprule

\textbf{AU} &\textbf{Guidance}  &AU    &ID  &BG   &HP  &\textbf{AU} &\textbf{Guidance}  &AU    &ID  &BG   &HP \\
 \cline{3-6} \cline{9-12}
\textbf{Dropout} &\textbf{Scale} & MSE ($\downarrow$)   & L2 ($\downarrow$)  & RMSE ($\downarrow$)  & RMSE ($\downarrow$) &\textbf{Dropout} &\textbf{Scale} & MSE ($\downarrow$)   & L2 ($\downarrow$)  & RMSE ($\downarrow$) & RMSE ($\downarrow$) \\

\hline

\ding{55} &-    & 0.360  &0.480    &0.044  &0.032                      &\ding{51} &6.5  & 0.293   &0.527   &0.050  &0.032\\  
\ding{51} &0.5  & 0.357  &0.482   &0.043  &0.032                       &\ding{51} &7.5  & 0.298    &0.516      &0.046    &0.034\\
\ding{51} &1.5  &0.333      &\textbf{0.473}     &0.043      &\textbf{0.031}              &\ding{51} &8.5  &0.305  &0.520   &0.046  &0.032\\ 
\ding{51} &2.5  &0.281      &\textbf{0.473}     &\textbf{0.042}  &0.032          &\ding{51} &9.5  &0.309  &0.504   &0.052  &\textbf{0.031}\\ 
\ding{51} &3.0  &\textbf{0.261}      &\textbf{0.473}   &0.044  &0.032 &\ding{51} &10.5 &0.315  &0.519   &0.055  &0.032\\ 
\ding{51} &3.5  &0.264      &0.484     &0.044  &0.032                  &\ding{51} &11.5 &0.323  &0.523   &0.052  &0.034\\ 
\ding{51} &4.5  &0.280      &0.495     &0.046  &0.034                  &\ding{51} &12.5 &0.331  &0.533   &0.057  &0.032\\ 
\ding{51} &5.5  &0.286      &0.510     &0.048  &0.034                  &\ding{51} &13.5 &0.350  &0.536   &0.055  &0.033\\

\hline
\end{tabular}
\end{table*}


\begin{table}[h!]
\centering
\caption{\textbf{Efficiency analysis of the training and inference stages of the proposed approach.} Per iter. means per iteration.} 
\label{tab:efficiency}
\begin{tabular}{lcccc}
\toprule
Setting & GPU Type & dtype & Memory & Time (s) \\

\midrule
Training   & A100 40GB ($\times 4$)     &FP32      &26.5GB  &23.6 (per iter.)   \\  
Inference   & A100 40GB      &FP32    &8.2GB   &8.4    \\
\bottomrule
\end{tabular}
\end{table}

\noindent \textbf{Efficiency Analysis.} The training and inference efficiency of our method is summarized in TABLE~\ref{tab:efficiency}. The results show that the proposed framework exhibits reasonable computational cost for a large-model fine-tuning task, although it is inherently slower than non-diffusion-based approaches. Nevertheless, this additional cost brings the benefit of substantially finer control over facial expressions and higher-quality editing results. Training on four A100-40GB GPUs requires 26.5~GB of memory and approximately 23.6 seconds per iteration under FP32 precision, which is acceptable given the scale of the model. During inference, the model operates on a single A100-40GB GPU with a memory footprint of only 8.2~GB and a runtime of 8.4 seconds per image. These results demonstrate that the method is practical to deploy and provides efficient performance at test time.

\section{Impact Statement}
The proposed MagicFace for facial expression editing offers diverse applications. It greatly enhances communication in digital environments by allowing individuals to convey themselves more effectively through avatars or digital characters. This improvement facilitates better interactions in virtual meetings, online education platforms, and social networking spaces. In addition, MagicFace has the potential to transform the entertainment and media industries by enabling the creation of more realistic and expressive characters in films, video games, and animations. This innovation enhances immersive storytelling and boosts audience engagement. Experiments show that our model can generalize across various real human ethnicities and age groups,  and even to out-of-domain images, e.g., cartoon-style images and painting-style images.

\noindent\textbf{Potential Negative Social Impact}: The method could be exploited for malicious purposes, such as creating fake animated images or videos of individuals, which might be used for fraudulent activities. To prevent such misuse, it is crucial to implement various measures, including digital watermarking and detection algorithms, enforcing strict legal regulations, promoting media literacy through public awareness and education, and establishing ethical guidelines within the tech industry. Achieving this requires collaborative efforts among technology companies, governments, educators, and the public to foster a safer digital environment and reduce the risks associated with fraudulent AI-generated content.

\section{Discussion and Conclusion}

\noindent\textbf{Limitations.} Our model may struggle to generate the same image completely when the AU variations are input as all zeros, partially because of the inaccuracy of the estimated AU intensity used in the training stage. Besides, our model demonstrates lower operational efficiency compared to non-diffusion-based methods due to the use of DDPM.

\noindent \textbf{Conclusion.} In this paper, we present \textit{MagicFace}, a framework capable of editing any portraits with different facial expressions and intensity levels by embedding action units as conditions into a Stable-Diffusion. We introduce an ID encoder that effectively retains detailed face attributes and achieves efficient facial expression controllability. Our approach provides a user-friendly way to enable fine-grained control over facial expressions with high quality, which outperforms existing face expression editing methods.  

\noindent\textbf{Future Work.} While the current work focuses on static AU-based facial expression editing, several promising extensions remain for future research. First, we aim to develop AU-based facial animation by enabling temporally consistent expression control across video frames. This is particularly important because in dynamic scenarios the head pose changes continuously, and the visible background consequently shifts from frame to frame. Addressing this requires incorporating a component that can infer and update the background as head pose and facial geometry evolve over time, thereby avoiding cross-frame inconsistencies and enabling smooth AU-driven facial motion. Second, we intend to scale our approach to high-resolution facial expression editing (e.g., 1024×1024 or higher). Achieving this level of fidelity will require more advanced generative backbones, such as SDXL \footnote{https://stablediffusionxl.com/} or other state-of-the-art diffusion architectures, which can better preserve identity and fine-grained facial details under AU manipulation. These two directions represent promising next steps toward more expressive and versatile facial expression editing.

\bibliographystyle{IEEEtran}
\bibliography{tmm_references}

\end{document}